\documentclass[11pt, letterpaper]{article}

\usepackage{booktabs}
\usepackage{microtype}
\usepackage[final]{acl}
\usepackage{graphicx}
\usepackage{subfigure}
\usepackage{subcaption}
\usepackage{placeins}
\usepackage{times}
\usepackage{mdframed}
\usepackage[export]{adjustbox}

\title{Query Timing Produces Opposite Positional Biases\\ Between LLMs and Humans}

\author{Jasin Cekinmez\textsuperscript{*}\hspace{1em} Addison J. Wu\textsuperscript{*}\hspace{1em} Thomas L. Griffiths \\
Princeton University}

\begin{document}

\maketitle

\begin{abstract}
Positional biases such as recency and primacy effects have been documented in large language models (LLMs), yet the underlying mechanism by which these models make their evaluations remains poorly understood. Both \emph{primacy} and \emph{recency} biases have been observed in human judgments in response to evidence, but recent work suggest that \emph{when} the listener updates their beliefs -- during the presentation of evidence or only at the end -- influences the presence of such effects. We investigate whether a similar phenomenon holds for LLMs, finding divergence from human behavior. These biases are more pronounced in newer models compared to their predecessors.
\end{abstract}

\section{Introduction}

Large language models are increasingly used not only to generate information, but also to make decisions on our behalf \cite{li2024llms, zheng2023judging}. As LLM-as-a-judge systems are adopted in more domains where judgments carry real consequences \cite{li2024llms}, assessing whether they can serve as fair and reliable evaluators becomes critical. Specifically, a growing body of literature has identified positional biases in LLM evaluations, where the order in which evidence is presented impacts the judgments made by the model \cite{schilcher2025characterizing, sun2025posum, wang2024large}. This literature adds to demonstrations of the brittleness of evaluative workflows assisted by LLMs-as-a-Judge, including stochasticity \cite{lee2025evaluating} and self-preferential bias \cite{panickssery2404llm}, flaws that are not reliably solved with simple interventions like fine-tuning \cite{zhu2023judgelm}. Such sensitivity to evidence order raises concerns about whether LLMs are relying on the substantive content of the evidence or on spurious structural details. While prior work establishes and provides insight into some of these biases, there is limited work fully delineating \emph{when} such specific biases affect LLM judgments. Understanding when these positional biases manifest in LLM judgments is a prerequisite for the responsible scaling of LLM-as-a-judge systems.

For human decision making, positional biases – effects in which the order of evidence is provided impacts judgments – are widely documented \cite{murdock1962serial}. However, previous results offer conflicting findings about what type of positional biases emerge in decision-making. Some results show \emph{primacy effects} in jury decisions \citep{dennis2001primacy, pennington1982witnesses, tetlock1983accountability, wells1985timing} – that is, in some studies, when a defendant presents their case first, they are less likely to be perceived as guilty. However, other studies have found \emph{recency effects} to be the prevailing mode of bias \citep{charman2016evidence, dahl2009investigating, furnham1986robustness, maegherman2022law}.

In an effort to reconcile such diverging conclusions, \citet{qiao2025speak} show that which positional biases manifest in humans is influenced by when people are asked for their momentary belief predictions. Two common response modes are asking for a final judgment at the end, known as End-of-Sequence (EoS) answering, and for intermediate judgments as updated evidence is presented, known as Step-by-Step (SbS) answering \citep{hogarth1992order, kerstholt1998judicial, qiao2025speak}.
\citet{qiao2025speak} found that people exhibit positional biases -- typically recency biases -- in the SbS response mode, but no overall bias in the EoS response mode. 

Building on this previous work, we examine order effects under both EoS and SbS response modes in LLM evaluation. We build on the paradigm introduced by \citet{qiao2025speak} which studies these effects in humans within a single accusatory setting. We extend this paradigm to three distinct accusatory settings to better understand whether these order effects and response mode effects generalize beyond the original setting. By doing so, we aim to assess whether the order effects observed in human cognition also manifest in LLM judgments across different contexts. Additionally, this design allows us to better understand the evolutionary nature of biases within LLMs, examining whether such biases remain stable or change across different models and successive model iterations.

\section{Methodology}

\subsection{Overview}
We investigate the impacts of different ways information is inputted to LLMs across multiple accusatory settings, adopting the experimental framework introduced by \citet{qiao2025speak}. We examine how SbS versus EoS response modes influence the final judgment made by the LLMs, as well as how the ordering of evidence within each response mode affects the final judgment. Our analysis spans both open and closed source models. We intentionally consider three separate accusatory domains to assess whether the trends observed by \citeauthor{qiao2025speak} generalize to other domains. Additionally, accusatory settings are utilized since conditional probability questions in these domains allow for an implicit causal link to the individual, making this an important domain to analyze how LLMs form their judgments.  

\subsection{Dataset and Prompts}
We employ three different accusatory settings for our experiments: criminal, academic, and social. The criminal setting involves a homicide and is borrowed from \citet{qiao2025speak}. The social misconduct setting involves deciding whether an employee caused damage to property, while the academic misconduct setting involves deciding whether a student committed academic dishonesty. The social and academic misconduct settings were designed to mimic the structure of \citet{qiao2025speak}. For all three settings, we include a neutral case summary which summarizes the situation, four pieces of prosecution evidence, and four pieces of defense evidence.

We evaluate LLM behavior using two types of questions. The first type of question conditions on the individual committing or not committing the action and then asks, based on that assumption, how likely it is that the stated piece(s) of evidence would occur. The first question type comes in sets of two, where the first question conditions on not being guilty and asks how likely that piece of evidence is, and the second question follows the same format but instead conditions on being guilty. The second type of question consists of final verdict questions. The first conditions on all of the evidence and asks how likely it is that the person is guilty, and the final question asks for the verdict for the person (Guilty/Not Guilty) (See Appendix A).

\subsection{Procedure}
For each LLM and each of the three accusatory settings, we run four experiments: SbS with prosecution evidence then defense evidence (PD), SbS (DP), EoS (PD), and EoS (DP). For each experiment, we perform 30 independent runs to account for variability in LLM judgments, and compute the proportion of guilty verdicts by dividing the number of guilty outcomes by 30.

The key distinction between SbS and EoS lies in how evidence is presented to the model. In the SbS condition, evidence is provided sequentially, the model is first presented with a single piece of evidence, followed by two conditional probability questions, before receiving the next piece of evidence. In contrast, in the EoS condition, all evidence from one side is presented at once, followed by the two questions, before proceeding to the opposing side.

In all experiments, the LLM is given a case summary. For both response modes, the pieces of evidence are randomly permuted within the prosecution and defense sets for each run. Evidence and intermediate questions are then given according to the assigned response mode and evidence order. After all evidence has been presented, the model is asked two final questions. The penultimate question asks how likely it is, given all the evidence, that the individual is guilty, and the final question asks whether the individual is guilty (Guilty/Not Guilty).

We conduct these experiments across both open and closed source LLMs, as well as across older and newer model versions, to examine whether response mode biases differ by model accessibility, release time, or model idiosyncrasies. 

\section{Results}


\begin{table}[ht!]
\centering
\caption{Verdict proportions for academic misconduct.}
\label{tab:academic_verdicts}

\resizebox{\linewidth}{!}{%
\begin{tabular}{lcccc}
\toprule
 & \multicolumn{2}{c}{EoS} & \multicolumn{2}{c}{SbS} \\
Model & DP & PD & DP & PD \\
\midrule
GPT 4o * & \textbf{1.00} & \textbf{0.033} & \textbf{0.30} & \textbf{0.033} \\
Claude 3.7 Sonnet * & \textbf{1.00} & \textbf{0.433} & 0.967 & 0.833 \\
Claude 4 Sonnet * & \textbf{1.00} & \textbf{0.20} & 1.00 & 0.967 \\
Gemini 2.5 Flash & 1.00 & 1.00 & 1.00 & 1.00 \\
Llama-4-Maverick & 1.00 & 0.8 & \textbf{0.6} & \textbf{0.033} \\
Qwen2.5-72b-Turbo * & \textbf{0.567} & \textbf{0.133} & 1.00 & 1.00 \\
\bottomrule
\end{tabular}
}

\caption*{\footnotesize
\textbf{Note.} \textbf{Bold} entries indicate statistically significant differences in the proportion of ``guilty'' verdicts within a prompting method.
\emph{Starred} * entries denote cases where EoS prompting results in more amplified biases than with SbS prompting.
}
\end{table}

\begin{table}[ht!]
\centering
\caption{Verdict proportions for social misconduct.}
\label{tab:social_verdicts}
\resizebox{\linewidth}{!}{%
\begin{tabular}{lcccc}
\toprule
 & \multicolumn{2}{c}{EoS} & \multicolumn{2}{c}{SbS} \\
Model & DP & PD & DP & PD \\
\midrule
GPT 4o * &       \textbf{0.2667} & \textbf{0.00} & 0.00 & 0.00 \\
Claude 3.7 Sonnet *         & \textbf{0.967} & \textbf{0.00} & 0.1 & 0.00 \\
Claude 4 Sonnet *        & \textbf{0.967} & \textbf{0.00} & \textbf{0.367} & \textbf{0.067} \\
Gemini 2.5 Flash         & 0.567 & 0.467 & 0.30 & 0.33 \\
Llama-4-Maverick *        & \textbf{0.967} & \textbf{0.00} & \textbf{0.167} & \textbf{0.00} \\
Qwen2.5-72b-Turbo *             & \textbf{0.9} & \textbf{0.033} & 0.367 & 0.167 \\
\bottomrule
\end{tabular}}
\end{table}

\begin{table}[ht!]
\centering
\caption{Verdict proportions for criminal misconduct.}
\label{tab:criminal_verdicts}
\resizebox{\linewidth}{!}{%
\begin{tabular}{lcccc}
\toprule
 & \multicolumn{2}{c}{EoS} & \multicolumn{2}{c}{SbS} \\
Model & DP & PD & DP & PD \\
\midrule
GPT 4o * &       \textbf{1.00} & \textbf{0.20} & 1.00 & 0.867 \\
Claude 3.7 Sonnet *       & \textbf{0.967} & \textbf{0.00} & 0.833 & 1.00 \\
Claude 4 Sonnet *        & \textbf{0.967} & \textbf{0.00} & 0.933 & 0.867 \\
Gemini 2.5 Flash         & 0.9 & 0.7 & 0.8 & 0.833 \\
Llama-4-Maverick *        & \textbf{0.867} & \textbf{0.033} & 0.8 & 0.933 \\
Qwen2.5-72B-Turbo               & 0.833 & 0.7 & 1.00 & 1.00 \\
Qwen2.5-7B-Turbo                & 1.00 & 1.00 & 1.00 & 1.00 \\
\bottomrule
\end{tabular}}
\end{table}

\subsection{General Trends}
\label{sec:general_trends}
We employ a two-stage Fisher’s Exact Test to assess statistical significance at the 0.05 level. First, we test whether the difference in proportions between the orders DP and PD is statistically significant within each response mode. If both order effects are statistically significant, we then test the difference of these differences between EoS and SbS. To account for multiple comparisons across models and settings, we apply a Benjamini--Hochberg correction; findings remain significant.

\citet{qiao2025speak} find that humans exhibit a recency bias under the SbS response mode, but no bias under the EoS response mode. However, we observe a contrasting pattern in LLMs. For most models, the order effect is not statistically significant in the SbS condition, but is statistically significant in the EoS condition, specifically, a recency bias. To rule out context length as a confound — and to better approximate human memory limits, since unlike LLMs humans cannot perfectly recall earlier evidence — we ran a compressed SbS ablation replacing prior responses with their numeric score. Results showed similarly minimal bias, suggesting SbS order effects are not an artifact of prompt length (see Appendix~\ref{sec:compress_appendix}). Additionally, even when both response mode order effects are statistically significant, the EoS order effects are still more statistically significantly pronounced in magnitude compared to the SbS order effects (Cohen's $h$ up to $\approx 2.1$ and odds ratios up to $\approx 260$; Tables~\ref{tab:academic_verdicts}, \ref{tab:social_verdicts}, \ref{tab:criminal_verdicts}). 

An exception to this trend is Gemini 2.5 Flash, which does not exhibit statistically significant order effects in either response mode. Additionally, order effects and response mode biases are less consistent for open source models than for GPT and Claude. Overall, the EoS recency bias holds robustly for GPT and Claude models across all three settings (Tables~\ref{tab:academic_verdicts}, \ref{tab:social_verdicts}, \ref{tab:criminal_verdicts}).

We additionally evaluate Gemini 3 Flash, a reasoning model that performs dynamic thinking by default (with a default thinking level of ``high''). It still exhibits a statistically significant EoS recency effect on the criminal setting ($0.37$ vs.\ $0.07$ guilty across evidence orders; Fisher's exact $p = .01$, Cohen's $h = 0.78$), while the one model that shows no order effect in our panel, Gemini 2.5 Flash, is itself also a thinking model. This finding suggests reasoning capability is therefore neither necessary nor sufficient for the bias.

\begin{figure}[ht!]
    \centering
    \includegraphics[width=0.75\linewidth]{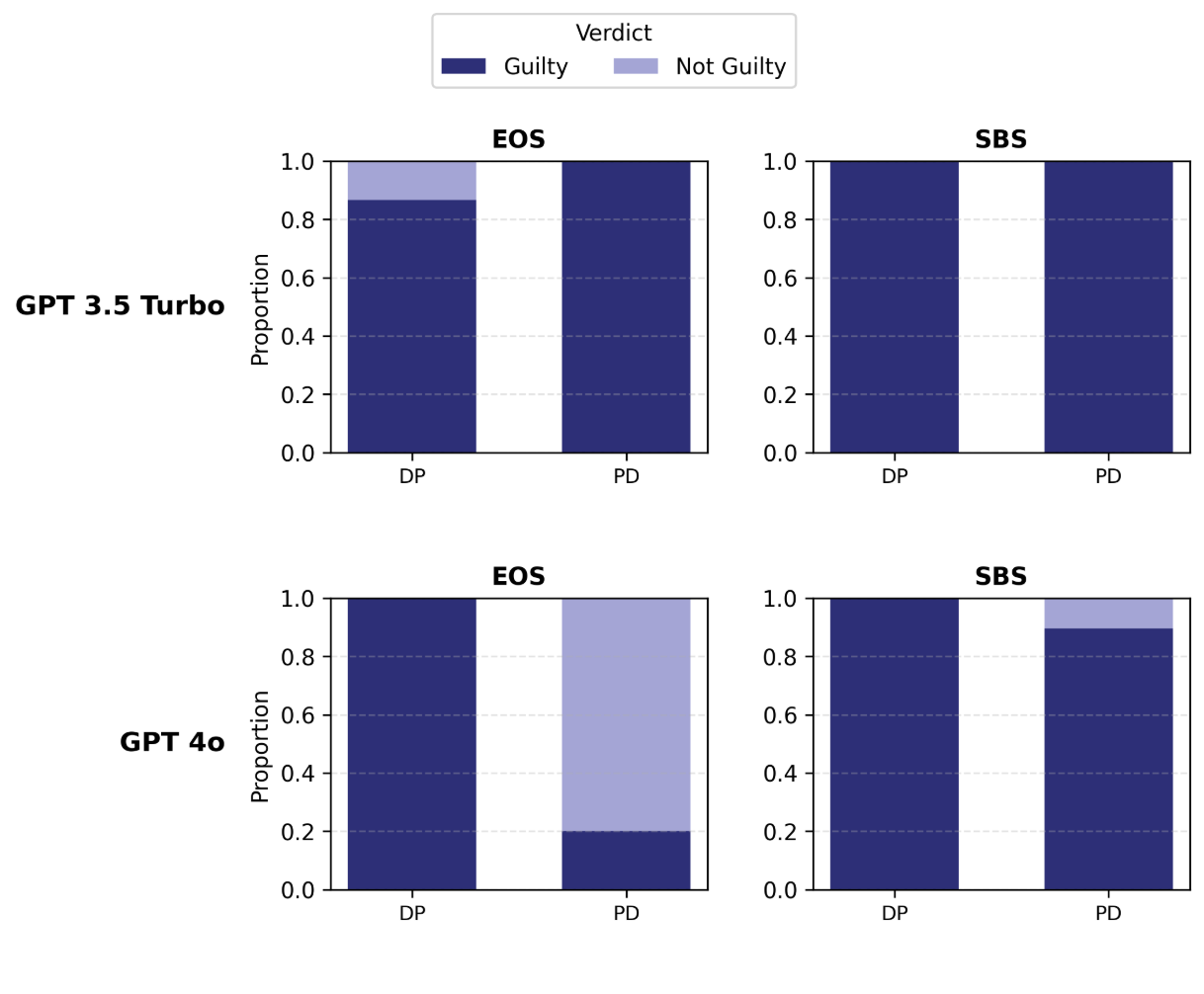}
    \caption{Changes in proportion of "guilty" verdicts across GPT model family for criminal misconduct setting}
    \label{fig:gpt_family}
\end{figure}

\begin{figure}[ht!]
    \centering
    \includegraphics[width=0.8\linewidth]{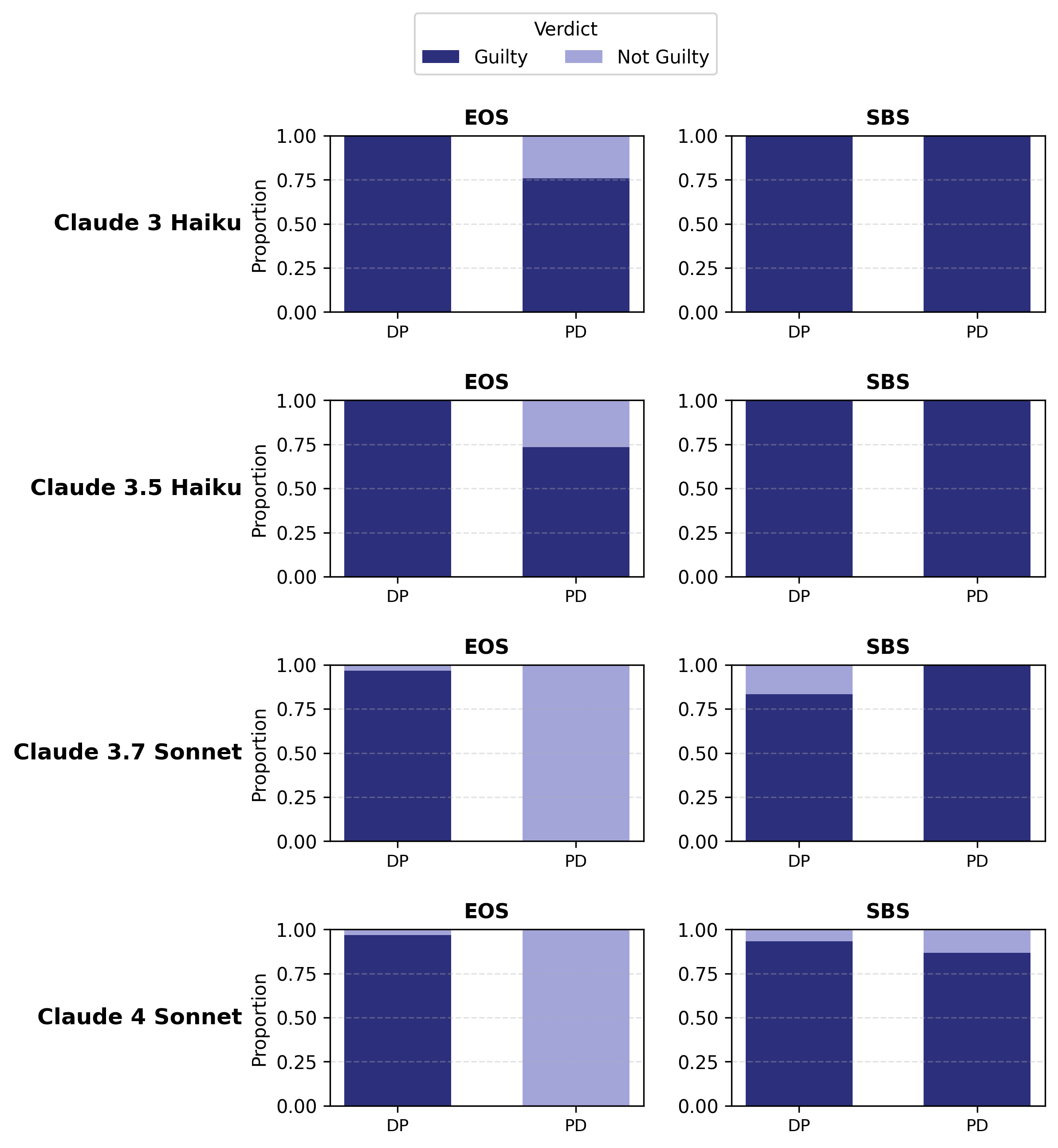}
    \caption{Changes in proportion of "guilty" verdicts across Claude models for criminal misconduct setting}
    \label{fig:claude_family}
\end{figure}

\subsection{Emergence of Biases Within Model Families}
As done in Section~\ref{sec:general_trends}, we employ the same statistical tests to assess whether differences between order effects are statistically significant. For GPT 3.5 Turbo, we find no statistically significant order effects in either response mode. In contrast, GPT 4o exhibits a statistically significant recency bias in the EoS response mode (Figure~\ref{fig:gpt_family}). A similar progression is observed for Claude models, where Claude 3 Haiku and Claude 3.5 Haiku show no statistically significant order effect in either response mode; however, successive models – Claude 3.7 Sonnet, Claude 4 Sonnet – both exhibit recency bias in the EoS response mode. In sum, these results suggest that order effect biases in LLMs are not static (Figure~\ref{fig:claude_family}). Instead, LLMs evolve from exhibiting no order effects to showing statistically significant order effects which may manifest in different directions and response modes.

A parallel pattern appears along model \emph{size} within a family. On the criminal setting, the smaller Qwen2.5-7B returns a guilty verdict in every order and response mode (Table~\ref{tab:criminal_verdicts}), exhibiting no order sensitivity, whereas the larger Qwen2.5-72B produces order-dependent verdicts that reach significance in the academic and social settings (Tables~\ref{tab:academic_verdicts}, \ref{tab:social_verdicts}). Because this comparison varies only scale within a fixed family and training pipeline, it isolates scale in a way that cross-model comparisons cannot: increasing scale does not remove the bias.  

\subsection{(Mis)-Correspondence with Internal Bayesian Probability Updates}

\begin{figure}[b!]
    \centering

    \subfigure[DP direction]{
        \includegraphics[width=0.62\linewidth]{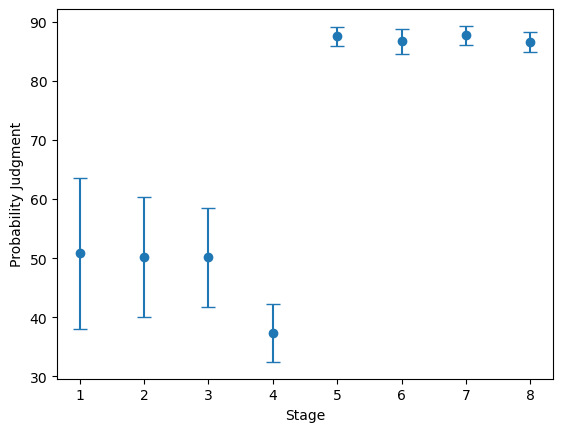}
        \label{fig:gpt_overtime}
    }


    \subfigure[PD direction]{
        \includegraphics[width=0.62\linewidth]{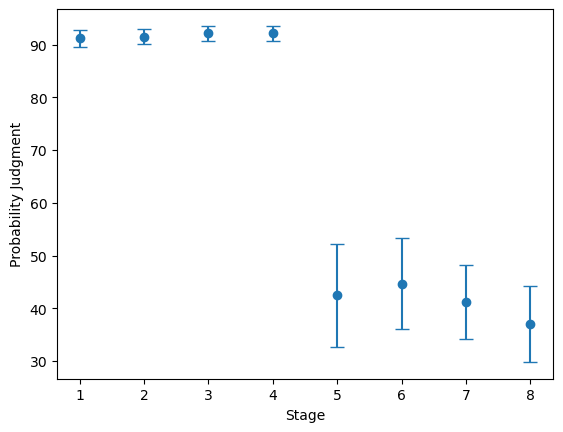}
        \label{fig:claude_overtime}
    }

    \caption{GPT-4o updated probability judgments of guilt for the Murder case in both SBS conditions}
    \label{fig:model_overtime}
\end{figure}

\noindent To understand why LLMs did not exhibit strong recency biases in their final verdict judgments in the SbS conditions, contrary to human participants, we analyzed the LLM-outputted probabilities of guilt after each piece of evidence was revealed. 

In both evidence orderings, GPT-4o frequently issued \emph{Not Guilty} final verdicts. However, this apparent stability in the final decision masked substantial intermediate belief updating. Specifically, the model’s reported probability of guilt systematically increased or decreased in response to individual evidence items, exhibiting clear sensitivity to both inculpatory and exculpatory information as it was sequentially presented (Figure~\ref{fig:model_overtime}).

Despite these intermediate belief shifts, changes in the direction or magnitude of the reported guilt probability were not reflected in the ultimate judgment, especially as indicated by the DP direction (Figure~\ref{fig:model_overtime}a), indicating a dissociation between belief reporting and decision making under SbS prompting. This goes against a simple thresholding account, in which the final verdict would be expected to  reflect the terminal probability estimate.  We replicate this trajectory analysis across all models with per-step verdict data in Appendix~\ref{sec:trajectory_appendix}, where the same pattern holds: intermediate guilt probabilities shift substantially and coherently with each evidence item, while final verdicts remain comparatively stable.

\section{Conclusion}
Overall, we find that LLMs exhibit positional biases that differ from those observed in human cognition. While humans tend to display recency effects in the SbS response mode, many LLMs instead exhibit recency biases under the EoS response mode. This pattern is strongest for GPT and Claude models, is less pronounced for open source models, and is absent for Gemini 2.5 Flash. Also, our results reveal a monotonic trend within model families: successive iterations consistently progress from exhibiting no statistically significant order effects to showing pronounced positional biases, as seen in both the GPT (3.5 $\to$ 4o) and Claude (3 $\to$ 3.5 Haiku $\to$ 3.7 $\to$ 4 Sonnet) lineages.

These findings highlight the need for caution when deploying LLMs as evaluators, particularly in high-stakes settings where judgments should be invariant to evidence order or response format. Although longer context evaluations may lead to exhibiting different biases, we do not pursue such scaling here, as the observed biases are already pronounced under the current experimental conditions.

For future directions, it is important to explore interventions at the level of response mode and evidence ordering that could yield more consistent judgments. Additionally, applying mechanistic interpretability techniques may help uncover the processes that lead to these biases, providing insight into why LLM evaluations diverge from human sequential reasoning in the first place.

\section*{Limitations}

Our study uses controlled accusatory scenarios with a fixed number and structure of evidence items, which allows for clear comparisons across response modes but may limit ecological validity. Real-world evaluative contexts often involve more complex, ambiguous, and interdependent evidence, and positional biases may manifest differently under such conditions. Additionally, while we evaluate multiple open- and closed-source LLM families, our model coverage is not exhaustive, and we do not isolate which aspects of model training or architecture drive the observed differences across model generations.

Moreover, our primary analyses focus on binary verdict outcomes, which compress richer belief states into coarse decisions. Although we examine intermediate probability judgments under Step-by-Step prompting, we treat these as behavioral outputs rather than direct readouts of internal states; the belief-updating/verdict dissociation we report holds at the level of the model's stated probabilities and verdicts regardless. Finally, while our response modes are adapted from human experimental paradigms, they may interact with LLM context processing in ways that lack a direct human analogue; we therefore frame the human--LLM contrast at the level of observed behavior under a matched query-timing manipulation, rather than as a claim about shared underlying mechanisms.

\bibliography{custom}

@inproceedings{sun2025posum,
  title={Po{S}um-Bench: Benchmarking Position Bias in {LLM}-based Conversational Summarization},
  author={Sun, Xu and Delphin-Poulat, Lionel and Tarnec, Christ{\`e}le and Shimorina, Anastasia},
  booktitle={Proceedings of the 2025 Conference on Empirical Methods in Natural Language Processing},
  year={2025}
}

@article{maegherman2022law,
  title={Law and order effects: {On} cognitive dissonance and belief perseverance},
  author={Maegherman, Enide and Ask, Karl and Horselenberg, Robert and Van Koppen, Peter J},
  journal={Psychiatry, Psychology and Law},
  volume={29},
  number={1},
  pages={33--52},
  year={2022},
  publisher={Taylor \& Francis}
}

@inproceedings{schilcher2025characterizing,
  title={Characterizing Positional Bias in Large Language Models: A Multi-Model Evaluation of Prompt Order Effects},
  author={Schilcher, Patrick and Karasin, Dominik and Sch{\"o}pf, Michael and Saleh, Haisam and Tommasel, Antonela and Schedl, Markus},
  booktitle={Findings of the Association for Computational Linguistics: EMNLP 2025},
  year={2025}
}

@article{kerstholt1998judicial,
  title={Judicial decision making: Order of evidence presentation and availability of background information},
  author={Kerstholt, Jos{\'e} H and Jackson, Janet L},
  journal={Applied Cognitive Psychology},
  volume={12},
  number={5},
  pages={445--454},
  year={1998},
  publisher={Wiley Online Library}
}

@article{hogarth1992order,
  title={Order effects in belief updating: The belief-adjustment model},
  author={Hogarth, Robin M and Einhorn, Hillel J},
  journal={Cognitive Psychology},
  volume={24},
  number={1},
  pages={1--55},
  year={1992},
  publisher={Elsevier}
}

@article{charman2016evidence,
  title={Evidence evaluation and evidence integration in legal decision-making: Order of evidence presentation as a moderator of context effects},
  author={Charman, Steve D and Carbone, Jon and Kekessie, Seyram and Villalba, Daniella K},
  journal={Applied Cognitive Psychology},
  volume={30},
  number={2},
  pages={214--225},
  year={2016},
  publisher={Wiley Online Library}
}

@article{dahl2009investigating,
  title={Investigating investigators: How presentation order influences participant--investigators’ interpretations of eyewitness identification and alibi evidence},
  author={Dahl, Leora C and Brimacombe, CA Elizabeth and Lindsay, D Stephen},
  journal={Law and Human Behavior},
  volume={33},
  number={5},
  pages={368--380},
  year={2009},
  publisher={Springer}
}

@article{furnham1986robustness,
  title={The robustness of the recency effect: Studies using legal evidence},
  author={Furnham, Adrian},
  journal={The Journal of General Psychology},
  volume={113},
  number={4},
  pages={351--357},
  year={1986},
  publisher={Taylor \& Francis}
}

@inproceedings{qiao2025speak,
  title={Speak Last and Step-by-Step: The Effect of Order and Response Mode on Evidence Evaluation},
  author={Qiao, Mengxuan Helen and Lagnado, David},
  booktitle={Proceedings of the 47th Annual Meeting of the Cognitive Science Society},
  year={2025}
}

@inproceedings{lee2025evaluating,
  title={Evaluating the consistency of {LLM} evaluators},
  author={Lee, Noah and Hong, Jiwoo and Thorne, James},
  booktitle={Proceedings of the 31st International Conference on Computational Linguistics},
  year={2025}
}

@article{li2024llms,
  title={{LLM}s-as-{Judges}: A comprehensive survey on {LLM}-based evaluation methods},
  author={Li, Haitao and Dong, Qian and Chen, Junjie and Su, Huixue and Zhou, Yujia and Ai, Qingyao and Ye, Ziyi and Liu, Yiqun},
  journal={arXiv preprint arXiv:2412.05579},
  year={2024}
}

@article{pennington1982witnesses,
  title={Witnesses and Their Testimony: Effects of Ordering on Juror Verdicts},
  author={Pennington, Donald C},
  journal={Journal of Applied Social Psychology},
  volume={12},
  number={4},
  pages={318--333},
  year={1982},
  publisher={Wiley Online Library}
}

@article{dennis2001primacy,
  title={Primacy in causal strength judgments: The effect of initial evidence for generative versus inhibitory relationships},
  author={Dennis, Martin J and Ahn, Woo-Kyoung},
  journal={Memory \& Cognition},
  volume={29},
  number={1},
  pages={152--164},
  year={2001},
  publisher={Springer}
}

@article{wells1985timing,
  title={The Timing of the Defense Opening Statement: Don't Wait Until the Evidence Is in},
  author={Wells, Gary L and Wrightsman, Lawrence S and Miene, Peter K},
  journal={Journal of Applied Social Psychology},
  volume={15},
  number={8},
  pages={758--772},
  year={1985},
  publisher={Wiley Online Library}
}

@article{tetlock1983accountability,
  title={Accountability and complexity of thought.},
  author={Tetlock, Philip E},
  journal={Journal of Personality and Social Psychology},
  volume={45},
  number={1},
  pages={74-83},
  year={1983},
  publisher={American Psychological Association}
}

@inproceedings{wang2024large,
  title={Large language models are not fair evaluators},
  author={Wang, Peiyi and Li, Lei and Chen, Liang and Cai, Zefan and Zhu, Dawei and Lin, Binghuai and Cao, Yunbo and Kong, Lingpeng and Liu, Qi and Liu, Tianyu and Sui, Zhifang},
  booktitle={Proceedings of the 62nd Annual Meeting of the Association for Computational Linguistics},
  year={2024}
}

@article{murdock1962serial,
  title={The serial position effect of free recall.},
  author={Murdock Jr, Bennet B},
  journal={Journal of Experimental Psychology},
  volume={64},
  number={5},
  pages={482-488},
  year={1962},
  publisher={American Psychological Association}
}

@article{zheng2023judging,
  title={Judging {LLM}-as-a-{Judge} with {MT}-{B}ench and {C}hatbot {A}rena},
  author={Lianmin Zheng and Wei-Lin Chiang and Ying Sheng and Siyuan Zhuang and Zhanghao Wu and Yonghao Zhuang and Zi Lin and Zhuohan Li and Dacheng Li and Eric P. Xing and Hao Zhang and Joseph E. Gonzalez and Ion Stoica},
  journal={Advances in Neural Information Processing Systems},
  volume={36},
  year={2023}
}

@article{
panickssery2404llm,
title={{LLM} Evaluators Recognize and Favor Their Own Generations},
author={Arjun Panickssery and Samuel R. Bowman and Shi Feng},
journal={Advances in Neural Information Processing Systems},
volume={38},
year={2024}
}

@inproceedings{
zhu2023judgelm,
title={Judge{LM}: Fine-tuned Large Language Models are Scalable Judges},
author={Lianghui Zhu and Xinggang Wang and Xinlong Wang},
booktitle={The Thirteenth International Conference on Learning Representations},
year={2025},
url={https://openreview.net/forum?id=xsELpEPn4A}
}

\newpage
\onecolumn
\appendix
\section{Prompts}

\subsection{Murder Case Setting}
\subsubsection{Case Summary Prompt}

\begin{figure}[ht!]
\centering
\fbox{%
\begin{minipage}{0.85\linewidth}
\ttfamily\small
Jane is charged with two counts of murder of both her children, and here is the background of the case:\\

After Jane and Frank had been happily married for 5 years, their first son, Andy, was born.\\

When Andy was 11 weeks old, Jane found the baby had fallen unconscious after being put to bed when she was alone with the baby at home. Jane soon called an ambulance, but Andy was declared dead after being transported to the hospital. A post-mortem was conducted and it suggested natural causes of death.\\

After one year, Jane and Frank’s second son, Ben, was born. When Ben was 8 weeks old,
Jane discovered that the baby was unwell, and the couple called an ambulance. However, Ben could not be resuscitated either and was pronounced dead. An autopsy was conducted and injuries to the ribs and spinal cord, hypoxic damage (low level of oxygen) to the brain, and haemorrhages to the eyes and eyelids were discovered.\\

As a result of this finding, further tests based on Andy’s (the first son) autopsy photographs were carried out, and found bruises to the arms and legs, a torn frenulum, and blood in the lung. Subsequently, Jane was charged with two counts of murder of both children by smothering them.\\

Another possible explanation for the death of Andy and Ben is sudden infant death syndrome (SIDS), also known as COT or CRIB death. This is a sudden and unexplainable death of infants during their first year of life. There are a number of risk factors that correlate with SIDS, but no clear mechanism or cause has been identified. The frequency of SIDS is 1 in 4,000 in the UK, and it is more common among boys than girls.
\end{minipage}
}
\caption{Initial case summary prompt for criminal murder case.}
\label{fig:criminal_case_summary}
\end{figure}

\subsubsection{Defense Evidence}

\begin{figure}[ht!]
\centering
\fbox{%
\begin{minipage}{0.85\linewidth}
\ttfamily\small

\textbf{(D1)} The injuries that were found on Ben were not noted by nurses when he was brought to the hospital.

\medskip
\textbf{(D2)} Despite the injuries found on Ben, no other typical injuries of a death due to smothering were found, such as petechiae (tiny red or purple spots that can appear on the skin).

\medskip
\textbf{(D3)} The spinal bleeding that was found in Ben is a common finding in natural death, and the dislocated ribs probably occurred post-mortem or due to resuscitation attempts.

\medskip
\textbf{(D4)} Andy’s injuries had initially been put down as due to the resuscitation attempts, and such injuries are consistent with attempts to revive a person, especially a baby.

\end{minipage}
}
\caption{Evidence pieces for the Defense side in the murder case. Each sub-entry corresponds to a distinct evidentiary claim.}
\label{fig:criminal_defense_evidence}
\end{figure}

\newpage
\subsubsection{Defense EOS Questions}

In the EOS setting, both of the questions in Figure \ref{fig:criminal_defense_eos} are prompted to the LLM immediately after all of the defense evidence in Figure \ref{fig:criminal_defense_evidence} is presented.

\begin{figure}[ht!]
\centering
\fbox{%
\begin{minipage}{0.85\linewidth}
\ttfamily\small
If Jane \textbf{\{is/is NOT\}} GUILTY, how likely is it that the injuries found on Ben were not present when he was brought to the hospital, the absence of injuries such as petechiae in Ben, the spinal bleeding and dislocated ribs found in Ben, and the injuries found in Andy?\end{minipage}
}
\caption{End-of-sequence (EOS) defense question, issued after all defense evidence is presented.}
\label{fig:criminal_defense_eos}
\end{figure}

\subsubsection{Defense SBS Questions}

In the SBS setting, the model is queried \emph{sequentially}, immediately after each individual piece of defense evidence is presented. Each question conditions only on the evidence observed up to that point, allowing us to isolate how belief updates occur when evidence is evaluated incrementally rather than in aggregate.

\begin{figure}[ht!]
\centering
\fbox{%
\begin{minipage}{0.85\linewidth}
\ttfamily\small

\textbf{After (D1):}  
If Jane \textbf{\{is/is NOT\}} GUILTY, how likely is it that the injuries found on Ben were not present when he was brought to the hospital?

\medskip
\textbf{After (D2):}  
If Jane \textbf{\{is/is NOT\}} GUILTY, how likely is the absence of injuries such as petechiae in Ben?

\medskip
\textbf{After (D3):}  
If Jane \textbf{\{is/is NOT\}} GUILTY, how likely is it that the spinal bleeding and dislocated ribs were found in Ben?

\medskip
\textbf{After (D4):}  
If Jane \textbf{\{is/is NOT\}} GUILTY, how likely are the injuries found in Andy?

\end{minipage}
}
\caption{Step-by-step (SBS) defense questions. Each query is issued immediately after the corresponding evidence item (D1–D4) is presented.}
\label{fig:criminal_defense_sbs}
\end{figure}

\subsubsection{Prosecution Evidence}

\begin{figure}[ht!]
\centering
\fbox{%
\begin{minipage}{0.85\linewidth}
\ttfamily\small

\textbf{(P1)} The similarities between both incidents are striking: both babies were about the same age when they died; both died at a similar time of day after being fed; and both were found unconscious by Jane when she was alone with them.

\medskip
\textbf{(P2)} A doctor alleged that the hypoxic damage to Ben’s brain must have been caused a matter of hours before death.

\medskip
\textbf{(P3)} The hypoxic damage to Ben’s brain and the haemorrhages to the eyes and eyelids are consistent with smothering and/or other violent trauma.

\medskip
\textbf{(P4)} The bruises to the arms and legs, the torn frenulum, and blood in the lung found in the re-examination of Andy’s death are indicative of abuse and characteristic of smothering.

\end{minipage}
}
\caption{Evidence pieces for the Prosecution side in the murder case. Each sub-entry corresponds to a distinct evidentiary claim.}
\label{fig:criminal_prosecution_evidence}
\end{figure}
\newpage
\subsubsection{Prosecution EOS Questions}

In the EOS setting, the question in Figure \ref{fig:criminal_prosecution_eos} is prompted to the LLM immediately after all of the prosecution evidence in Figure \ref{fig:criminal_prosecution_evidence} is presented.

\begin{figure}[ht!]
\centering
\fbox{%
\begin{minipage}{0.85\linewidth}
\ttfamily\small
If Jane \textbf{\{is/is NOT\}} GUILTY, how likely are there to be these similarities between both incidents, the hypoxic damage to Ben’s brain was caused a matter of hours before death, the hypoxic damage to Ben’s brain and haemorrhages to the eyes and eyelids, and the bruises to the arms and legs, the torn frenulum, and blood in the lung in Andy?
\end{minipage}
}
\caption{End-of-sequence (EOS) prosecution question, issued after all prosecution evidence is presented.}
\label{fig:criminal_prosecution_eos}
\end{figure}

\subsubsection{Prosecution SBS Questions}

In the SBS setting, the model is queried sequentially, immediately after each individual piece of prosecution evidence is presented. Each query conditions only on the evidence observed up to that point, enabling analysis of incremental belief updating rather than end-of-sequence aggregation.

\begin{figure}[ht!]
\centering
\fbox{%
\begin{minipage}{0.85\linewidth}
\ttfamily\small

\textbf{After (P1):}  
If Jane \textbf{\{is/is NOT\}} GUILTY, how likely are there to be similarities between both incidents?

\medskip
\textbf{After (P2):}  
If Jane \textbf{\{is/is NOT\}} GUILTY, how likely is it that the hypoxic damage to Ben’s brain was caused a matter of hours before death?

\medskip
\textbf{After (P3):}  
If Jane \textbf{\{is/is NOT\}} GUILTY, how likely are the hypoxic damage to Ben’s brain and the haemorrhages to the eyes and eyelids?

\medskip
\textbf{After (P4):}  
If Jane \textbf{\{is/is NOT\}} GUILTY, how likely are the bruises to the arms and legs, the torn frenulum, and the presence of blood in the lung in Andy?

\end{minipage}
}
\caption{Step-by-step (SBS) prosecution questions. Each query is issued immediately after the corresponding evidence item (P1–P4) is presented.}
\label{fig:criminal_prosecution_sbs}
\end{figure}

\subsubsection{Verdict Elicitation Prompt}

\begin{figure}[ht!]
\centering
\fbox{%
\begin{minipage}{0.85\linewidth}
\ttfamily\small

Please state your verdict for the defendant (Guilty / Not guilty).

\end{minipage}
}
\caption{Prompt used to elicit the model’s final verdict.}
\label{fig:criminal_verdict}
\end{figure}

\newpage
\subsection{Academic Misconduct Setting}
\subsubsection{Case Summary Prompt}

\begin{figure}[ht!]
\centering
\fbox{%
\begin{minipage}{0.85\linewidth}
\ttfamily\small
Lena is a third-year PhD student in a cognitive neuroscience lab. She is working on a high-profile experiment testing a new theory about memory. The study involves 80 participants, and her supervisor, Professor Mills, is under pressure to produce publishable findings before a major grant review.\\

Lena runs the study and processes the data herself. The first draft of the paper reports strong, statistically significant support for the theory: the main effect is in the predicted direction with $p = .01$, and several secondary analyses also yield clean results. The paper is submitted to a top journal and receives enthusiastic initial reviews.\\

However, concerns are raised after an anonymous whistleblower emails the department, claiming that some of Lena’s data look unusually consistent and that key analyses were changed after reviewers’ comments. An internal inquiry is opened into possible research misconduct, including fabrication or falsification of data.\\

During the inquiry, multiple pieces of information emerge. Some patterns in the dataset and analysis history appear suspicious, but there are also plausible alternative explanations: the lab recently switched analysis software; some files are missing due to a hard drive failure; and Lena has a previously strong record of careful data handling.\\

Lena maintains that any irregularities are the result of honest mistakes or miscommunication rather than intentional misconduct. The evaluation centers on whether Lena deliberately manipulated or fabricated data in this study, as opposed to the results being genuine and any errors being unintentional.
\end{minipage}
}
\caption{Initial case summary prompt for the academic misconduct scenario.}
\label{fig:academic_case_summary}
\end{figure}

\subsubsection{Academic Misconduct Defense Evidence}

\begin{figure}[ht!]
\centering
\fbox{%
\begin{minipage}{0.85\linewidth}
\ttfamily\small

\textbf{(D1)} The lab’s data management logs show that Lena ran and stored multiple alternative analysis scripts with varying exclusion criteria as part of a lab-wide effort to assess robustness. Other students in the lab exhibit similar patterns in their analysis scripts.

\medskip
\textbf{(D2)} Two participants in the original dataset reported falling asleep during the task or not understanding the instructions. Both participants were excluded based on notes in the raw data files, and these notes predate the journal submission.

\medskip
\textbf{(D3)} The lab switched from one statistical software package to another midway through the project, and several files were lost following a hard drive failure on Lena’s laptop. The IT department confirms that a hardware failure was reported and repaired during this period.

\medskip
\textbf{(D4)} Lena has no prior record of data irregularities. Her previous studies in the same lab passed an independent data audit conducted the year before, and her raw data and scripts were rated as well-documented and reproducible.

\end{minipage}
}
\caption{Defense evidence for the academic misconduct case. Each sub-entry corresponds to a distinct evidentiary claim.}
\label{fig:academic_defense_evidence}
\end{figure}

\newpage
\subsubsection{Academic Defense EOS Questions}

In the EOS setting, the questions in Figure \ref{fig:academic_defense_eos} are prompted to the LLM immediately after all defense evidence in Figure \ref{fig:academic_defense_evidence} is presented.

\begin{figure}[ht!]
\centering
\fbox{%
\begin{minipage}{0.85\linewidth}
\ttfamily\small
If Lena \textbf{\{did/did NOT\}} commit academic misconduct, how likely is it that the lab’s data management logs would show her running and storing multiple alternative analysis scripts with different exclusion criteria similar to other students, that two participants would be excluded based on notes predating the journal submission, that the lab would switch software mid-project and her laptop would suffer a confirmed hard drive failure leading to missing files, and that her previous studies would have passed an independent data audit and been rated as well-documented and reproducible?
\end{minipage}
}
\caption{End-of-sequence (EOS) defense question for the academic misconduct case.}
\label{fig:academic_defense_eos}
\end{figure}

\subsubsection{Academic Defense SBS Questions}

In the SBS setting, the model is queried sequentially, immediately after each individual piece of defense evidence is presented. Each query conditions only on the evidence observed up to that point, allowing isolation of incremental belief updating.

\begin{figure}[ht!]
\centering
\fbox{%
\begin{minipage}{0.85\linewidth}
\ttfamily\small

\textbf{After (D1):}  
If Lena \textbf{\{did/did NOT\}}  commit academic misconduct, how likely is it that the lab’s data management logs would show her running and storing multiple alternative analysis scripts with different exclusion criteria, similar to other students?

\medskip
\textbf{After (D2):}  
If Lena \textbf{\{did/did NOT\}}  commit academic misconduct, how likely is it that two participants who reported falling asleep or not understanding the instructions would be excluded based on notes that predate the journal submission?

\medskip
\textbf{After (D3):}  
If Lena \textbf{\{did/did NOT\}}  commit academic misconduct, how likely is it that the lab would switch software during the project and that Lena’s laptop would suffer a confirmed hard drive failure, leading to some missing files?

\medskip
\textbf{After (D4):}  
If Lena \textbf{\{did/did NOT\}}  commit academic misconduct, how likely is it that her previous studies would have passed an independent data audit and been rated as well-documented and reproducible?

\end{minipage}
}
\caption{Step-by-step (SBS) defense questions for the academic misconduct case. Each query is issued immediately after the corresponding evidence item (D1–D4) is presented.}
\label{fig:academic_defense_sbs}
\end{figure}

\newpage
\subsubsection{Academic Prosecution Evidence}

\begin{figure}[ht!]
\centering
\fbox{%
\begin{minipage}{0.85\linewidth}
\ttfamily\small

\textbf{(P1)} The final dataset exhibits an unusually high concentration of $p$-values just below the conventional significance threshold (e.g., $.047$, $.049$, $.041$) across several exploratory analyses, with very few $p$-values just above $.05$.

\medskip
\textbf{(P2)} File timestamps indicate that the preregistration document was modified after data collection had been completed, altering some planned exclusion criteria and primary outcomes.

\medskip
\textbf{(P3)} An earlier version of the analysis script, recovered from an automated backup, includes additional participants who were subsequently excluded from the final dataset.

\medskip
\textbf{(P4)} In private emails to a friend, Lena wrote that her supervisor was counting on the study to secure a grant, that a null result would severely harm her postdoctoral prospects, and that she needed the study to succeed.

\end{minipage}
}
\caption{Prosecution evidence for the academic misconduct case. Each sub-entry corresponds to a distinct evidentiary claim.}
\label{fig:academic_prosecution_evidence}
\end{figure}

\subsubsection{Academic Prosecution EOS Questions}

In the EOS setting, the questions in Figure \ref{fig:academic_prosecution_eos} are prompted to the LLM immediately after all prosecution evidence in Figure \ref{fig:academic_prosecution_evidence} is presented.

\begin{figure}[ht!]
\centering
\fbox{%
\begin{minipage}{0.85\linewidth}
\ttfamily\small
If Lena \textbf{\{did/did NOT\}} commit academic misconduct, how likely is it that the final dataset would show an unusually high concentration of $p$-values just below $.05$ and very few just above $.05$, that the preregistration document would be modified after data collection to change exclusion criteria and primary outcomes, that an earlier analysis script would include participants who were later excluded from the final dataset, and that Lena would write emails stating that her supervisor was counting on the study, that a null result would harm her career prospects, and that she needed the study to succeed?
\end{minipage}
}
\caption{End-of-sequence (EOS) prosecution question for the academic misconduct case.}
\label{fig:academic_prosecution_eos}
\end{figure}

\subsubsection{Academic Prosecution SBS Questions}

In the SBS setting, the model is queried sequentially, immediately after each individual piece of prosecution evidence is presented. Each query conditions only on the evidence observed up to that point, enabling analysis of incremental belief updating.

\begin{figure}[ht!]
\centering
\fbox{%
\begin{minipage}{0.85\linewidth}
\ttfamily\small

\textbf{After (P1):}  
If Lena \textbf{\{did/did NOT\}} commit academic misconduct, how likely is it that the final dataset would show an unusually high concentration of $p$-values just below $.05$ and very few just above $.05$ across multiple analyses?

\medskip
\textbf{After (P2):}  
If Lena \textbf{\{did/did NOT\}} commit academic misconduct, how likely is it that the preregistration document would be modified after data collection to change some exclusion criteria and primary outcomes?

\medskip
\textbf{After (P3):}  
If Lena \textbf{\{did/did NOT\}} commit academic misconduct, how likely is it that an earlier analysis script would include participants who were later excluded from the final dataset?

\medskip
\textbf{After (P4):}  
If Lena \textbf{\{did/did NOT\}} commit academic misconduct, how likely is it that she would write emails stating that her supervisor was counting on the study, that a null result would hurt her postdoctoral prospects, and that she needed the study to succeed?

\end{minipage}
}
\caption{Step-by-step (SBS) prosecution questions for the academic misconduct case. Each query is issued immediately after the corresponding evidence item (P1–P4) is presented.}
\label{fig:academic_prosecution_sbs}
\end{figure}

\subsubsection{Verdict Elicitation Prompt}

\begin{figure}[ht!]
\centering
\fbox{%
\begin{minipage}{0.85\linewidth}
\ttfamily\small
Please state whether you think Lena committed academic misconduct (Yes / No).
\end{minipage}
}
\caption{Prompt used to elicit the model’s final verdict in the academic misconduct case.}
\label{fig:academic_verdict}
\end{figure}
\newpage
\subsection{Vandalism Case Setting}
\subsubsection{Case Summary Prompt}

\begin{figure}[ht!]
\centering
\fbox{%
\begin{minipage}{0.85\linewidth}
\ttfamily\small
In a small open-plan office, the only shared printer suddenly stops working late in the afternoon. When employees attempt to print, the machine displays a critical hardware error and will not restart. The side panel is slightly ajar, and the paper tray is cracked.\\

Alex, a mid-level employee who frequently uses this printer, becomes the focus of informal suspicion. Earlier in the day, Alex had complained about the printer repeatedly jamming, and at least one coworker recalls seeing Alex near the printer shortly before it stopped working. Later, small plastic fragments are found in the trash can beneath Alex’s desk.\\

However, additional information complicates the situation. Building access logs show that Alex badged out of the office shortly before the time of the critical error recorded in the printer log, though the logs do not indicate the condition of the printer when Alex last used or approached it. Security footage also shows another coworker, Jamie, tugging forcefully on a jammed sheet of paper in the printer earlier that afternoon, but does not capture the final moments before the failure occurred.\\

The building technician notes that this printer model has a known history of failure due to a defective fuser unit. Alex has also previously reported issues with office equipment. These background facts do not, on their own, establish whether the printer malfunction resulted from ordinary use, rough handling by one of the employees, or an underlying mechanical defect.\\

Alex is being evaluated for whether their actions caused the printer to break (for example, through reckless or intentional handling), as opposed to the breakdown being attributable to ordinary use, another individual’s actions, or an underlying mechanical failure.
\end{minipage}
}
\caption{Initial case summary prompt for the vandalism scenario.}
\label{fig:vandalism_case_summary}
\end{figure}

\subsubsection{Vandalism Defense Evidence}

\begin{figure}[ht!]
\centering
\fbox{%
\begin{minipage}{0.85\linewidth}
\ttfamily\small

\textbf{(D1)} Building access logs show that Alex badged out of the office at 5:12 pm. The printer’s internal error log indicates that the critical hardware error occurred at 5:19 pm, seven minutes after Alex had left the building.

\medskip
\textbf{(D2)} Security camera footage from earlier that afternoon shows another coworker, Jamie, pulling repeatedly and somewhat forcefully on a jammed sheet of paper in the printer. Jamie is the last clearly visible person using the printer on the footage prior to the failure.

\medskip
\textbf{(D3)} The building technician reports that this printer model has a documented issue with its fuser unit. The same model has failed in a similar manner in two other departments within the past six months, in both cases without evidence of improper handling.

\medskip
\textbf{(D4)} Three days before the breakdown, Alex emailed facilities reporting that the printer was making a grinding noise and might require maintenance. In prior incidents involving broken office equipment, Alex has consistently reported problems promptly and cooperated with maintenance staff.

\end{minipage}
}
\caption{Defense evidence for the vandalism case. Each sub-entry corresponds to a distinct evidentiary claim.}
\label{fig:vandalism_defense_evidence}
\end{figure}
\newpage
\subsubsection{Vandalism Defense EOS Questions}

In the EOS setting, the questions in Figure \ref{fig:vandalism_defense_eos} are prompted to the LLM immediately after all defense evidence in Figure \ref{fig:vandalism_defense_evidence} is presented.

\begin{figure}[ht!]
\centering
\fbox{%
\begin{minipage}{0.85\linewidth}
\ttfamily\small
If Alex \textbf{\{did/did NOT\}} break the printer, how likely is it that he would have badged out of the building seven minutes before the critical error recorded in the printer’s log, that security camera footage would show Jamie pulling repeatedly and somewhat forcefully on a jammed sheet of paper earlier that afternoon, that the printer model would have a documented history of failing in the same way in other departments, and that Alex would have previously emailed facilities about a grinding noise and shown a pattern of promptly reporting equipment problems?
\end{minipage}
}
\caption{End-of-sequence (EOS) defense question for the vandalism case.}
\label{fig:vandalism_defense_eos}
\end{figure}

\subsubsection{Vandalism Defense SBS Questions}

In the SBS setting, the model is queried sequentially, immediately after each individual piece of defense evidence is presented. Each query conditions only on the evidence observed up to that point, allowing analysis of incremental belief updating.

\begin{figure}[ht!]
\centering
\fbox{%
\begin{minipage}{0.85\linewidth}
\ttfamily\small

\textbf{After (D1):}  
If Alex \textbf{\{did/did NOT\}} break the printer, how likely is it that he badged out of the building seven minutes before the critical error recorded in the printer’s log?

\medskip
\textbf{After (D2):}  
If Alex \textbf{\{did/did NOT\}} break the printer, how likely is it that security camera footage would show Jamie pulling repeatedly and somewhat forcefully on a jammed sheet of paper in the printer earlier that afternoon?

\medskip
\textbf{After (D3):}  
If Alex \textbf{\{did/did NOT\}} break the printer, how likely is it that this printer model would have a documented history of failing in the same way in other departments without evidence of improper handling?

\medskip
\textbf{After (D4):}  
If Alex \textbf{\{did/did NOT\}} break the printer, how likely is it that he would have emailed facilities about a grinding noise days earlier and have a prior pattern of promptly reporting broken equipment?

\end{minipage}
}
\caption{Step-by-step (SBS) defense questions for the vandalism case. Each query is issued immediately after the corresponding evidence item (D1–D4) is presented.}
\label{fig:vandalism_defense_sbs}
\end{figure}

\subsubsection{Vandalism Prosecution Evidence}

\begin{figure}[ht!]
\centering
\fbox{%
\begin{minipage}{0.85\linewidth}
\ttfamily\small

\textbf{(P1)} A coworker reports seeing Alex standing directly next to the printer approximately five minutes before it stopped working, and did not observe anyone else approach the printer during that interval.

\medskip
\textbf{(P2)} Earlier that day, Alex posted in the team chat that the printer had jammed again and stated, “If it does this one more time I’m going to lose it.” Several coworkers reacted with laughing emojis.

\medskip
\textbf{(P3)} After the printer failure, facilities staff found small plastic fragments in the trash can under Alex’s desk. The technician notes that the fragments appear similar in color and texture to the broken edge of the printer’s paper tray, though an exact match was not confirmed.

\medskip
\textbf{(P4)} When first asked informally about the printer, Alex stated that they had not used it that day. In a later conversation, Alex revised their statement, saying that they might have printed something quickly earlier in the morning but were not certain.

\end{minipage}
}
\caption{Prosecution evidence for the vandalism case. Each sub-entry corresponds to a distinct evidentiary claim.}
\label{fig:vandalism_prosecution_evidence}
\end{figure}

\subsubsection{Vandalism Prosecution EOS Questions}

In the EOS setting, the questions in Figure \ref{fig:vandalism_prosecution_eos} are prompted to the LLM immediately after all prosecution evidence in Figure \ref{fig:vandalism_prosecution_evidence} is presented.

\begin{figure}[ht!]
\centering
\fbox{%
\begin{minipage}{0.85\linewidth}
\ttfamily\small
If Alex \textbf{\{did/did NOT\}} break the printer, how likely is it that a coworker would see him standing next to the printer about five minutes before it stopped working with no one else approaching in that interval, that he would post in the team chat complaining about the printer and stating he was going to lose it if it jammed again, that plastic fragments similar in appearance to the broken tray would be found in the trash can under his desk, and that he would initially deny using the printer that day before later saying he might have printed something that morning?
\end{minipage}
}
\caption{End-of-sequence (EOS) prosecution question for the vandalism case.}
\label{fig:vandalism_prosecution_eos}
\end{figure}

\subsubsection{Vandalism Prosecution SBS Questions}

In the SBS setting, the model is queried sequentially, immediately after each individual piece of prosecution evidence is presented. Each query conditions only on the evidence observed up to that point, enabling analysis of incremental belief updating.

\begin{figure}[ht!]
\centering
\fbox{%
\begin{minipage}{0.85\linewidth}
\ttfamily\small

\textbf{After (P1):}  
If Alex \textbf{\{did/did NOT\}} break the printer, how likely is it that a coworker would see him standing next to the printer about five minutes before it stopped working, with no one else approaching in that interval?

\medskip
\textbf{After (P2):}  
If Alex \textbf{\{did/did NOT\}} break the printer, how likely is it that he would have posted in the team chat complaining about the printer jamming and stating that he was going to lose it if it jammed again?

\medskip
\textbf{After (P3):}  
If Alex \textbf{\{did/did NOT\}} break the printer, how likely is it that plastic fragments similar in appearance to the broken tray would be found in the trash can under his desk?

\medskip
\textbf{After (P4):}  
If Alex \textbf{\{did/did NOT\}} break the printer, how likely is it that he would initially deny using the printer that day and later say that he might have printed something that morning?

\end{minipage}
}
\caption{Step-by-step (SBS) prosecution questions for the vandalism case. Each query is issued immediately after the corresponding evidence item (P1–P4) is presented.}
\label{fig:vandalism_prosecution_sbs}
\end{figure}

\subsubsection{Verdict Elicitation Prompt}

\begin{figure}[ht!]
\centering
\fbox{%
\begin{minipage}{0.85\linewidth}
\ttfamily\small
Please state whether you think Alex broke the printer (Yes / No).
\end{minipage}
}
\caption{Prompt used to elicit the model’s final verdict in the vandalism case.}
\label{fig:vandalism_verdict}
\end{figure}

\newpage
\section{SbS-Compressed Ablation Results}
\label{sec:compress_appendix}

In the compressed SbS condition, after every two evidence blocks, all prior model responses are replaced with their trailing numeric score. This reduces accumulated context length while preserving the sequential query structure, serving as an ablation for the confound between response mode and prompt architecture, and more closely approximating human memory limitations by degrading access to earlier evidence. Table~\ref{tab:compress_verdicts} reports verdict proportions for the criminal misconduct setting under this condition.

\begin{table}[ht!]
\centering
\caption{Verdict proportions for criminal misconduct under SbS-Compressed condition.}
\label{tab:compress_verdicts}
\begin{tabular}{lcc}
\toprule
Model & DP & PD \\
\midrule
GPT 4o          & 0.900 & 0.900 \\
Gemini 2.5 Flash & 0.667 & 0.600 \\
Claude 4 Sonnet (4-6)          & 0.033 & 0.233 \\
Claude 4 Sonnet (4-20250514)   & 1.000 & 0.400 \\
Meta Llama      & 0.133 & 0.100 \\
Qwen            & 1.000 & 1.000 \\
\bottomrule
\end{tabular}
\caption*{\footnotesize \textbf{Note.} Proportions computed over 30 runs. No model shows a statistically significant shift in bias pattern relative to the standard SbS condition, consistent with the main finding that SbS order effects are not attributable to context length or prompt architecture.}
\end{table}

\clearpage
\section{Guilt-Probability Trajectories Across Models}
\label{sec:trajectory_appendix}

\begin{figure}[ht!]
\centering
\includegraphics[width=0.72\linewidth]{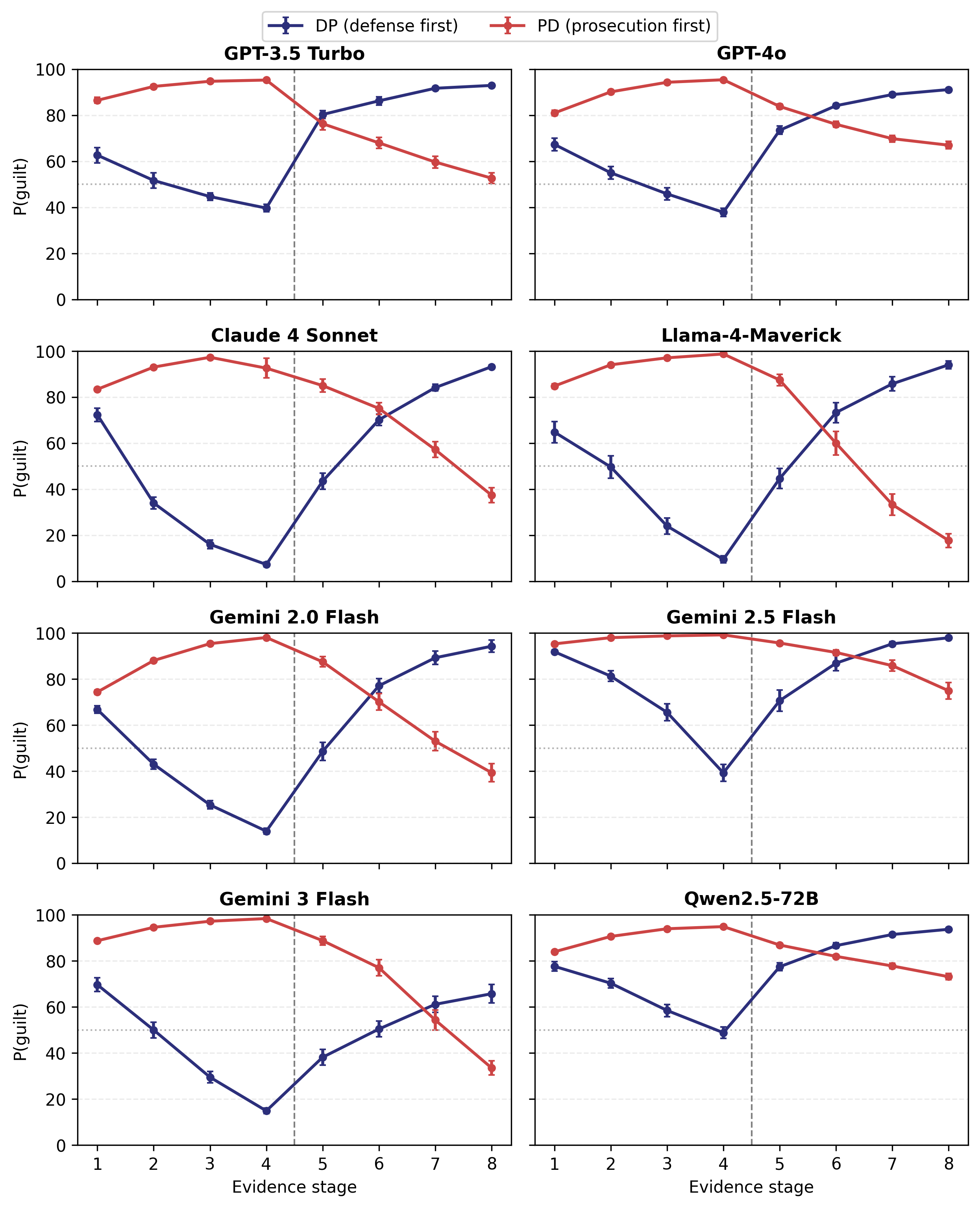}
\caption{Reported probability of guilt after each evidence item under Step-by-Step prompting, for all models with per-step verdict data on the criminal setting. Each panel shows the two evidence orders (DP: defense presented first; PD: prosecution presented first), averaged over 30 runs; error bars denote $\pm 1$ SEM. The dashed vertical line marks the switch between the first and second block of evidence (stage 4 to 5); the dotted horizontal line marks indifference ($50$).}
\label{fig:trajectories_appendix}
\end{figure}

Figure~\ref{fig:model_overtime} in the main text illustrates the belief-updating/verdict dissociation for a single model (GPT-4o). Figure~\ref{fig:trajectories_appendix} extends this analysis to all models with per-step verdict data on the criminal setting, plotting the model's reported probability of guilt after each evidence item (averaged over 30 runs) under both evidence orders. Across models, intermediate guilt probabilities shift substantially and coherently with the evidence---rising as inculpatory items accrue and falling as exculpatory items accrue---and the DP and PD trajectories mirror each other around the point where the evidence side switches. This continuous view reproduces, across the panel, the same dissociation reported for the binary verdicts: intermediate beliefs move with the evidence even where the final verdict is comparatively stable.

\end{document}